\documentclass[journal]{IEEEtran}
\usepackage{framed}
\usepackage[ruled]{algorithm2e}
\usepackage{amssymb}
\usepackage{latexsym}
\usepackage{epsfig}
\usepackage{graphicx}
\usepackage{booktabs,multirow, multicol}
\usepackage{cite}
\usepackage{epsfig}
\usepackage{caption}
 
\usepackage{epstopdf}

\usepackage{url}
\usepackage{xcolor}
\definecolor{newcolor}{rgb}{.8,.349,.1}

\ifCLASSINFOpdf

\fi

\begin{document}

\title{Fast and Accurate Algorithm for Eye Localization for Gaze Tracking in
Low Resolution Images}

\author{Anjith~George,~\IEEEmembership{Member,~IEEE,}
        and~Aurobinda~Routray,~\IEEEmembership{Member,~IEEE}
\thanks{ This paper is a preprint of a paper accepted by IET Computer Vision and is subject to Institution of Engineering and Technology Copyright. When the final version is published, the copy of record will be available at IET Digital Library. The proofed version is available in: Anjith George and Aurobinda Routray. "Fast and Accurate Algorithm for Eye Localization for Gaze Tracking in
Low Resolution Images, IET Computer Vision, 2016, DOI: http://dx.doi.org/10.1049/iet-cvi.2015.0316.}}

\maketitle

\begin{abstract}
Iris centre localization in low-resolution visible images is a challenging problem in
computer vision community due to noise, shadows, occlusions, pose variations, eye blinks, etc. This
paper proposes an efficient method for determining iris centre in low-resolution images in the visible
spectrum. Even low-cost consumer-grade webcams can be used for gaze tracking without any
additional hardware. A two-stage algorithm is proposed for iris centre localization. The proposed
method uses geometrical characteristics of the eye. In the first stage, a fast convolution based
approach is used for obtaining the coarse location of iris centre (IC). The IC location is further
refined in the second stage using boundary tracing and ellipse fitting. The algorithm has been
evaluated in public databases like BioID, Gi4E and is found to outperform the state of the art
methods.
\end{abstract}
\IEEEpeerreviewmaketitle

\section{Introduction}

Real-time detection and tracking of the eye is an active area of research in computer vision community. Localization and tracking of the eye can be useful in face alignment, gaze tracking, human-computer interaction ~\cite{hansen2010eye}, etc. The methods available to find iris position are based on 1) scleral coil 2) Electro oculography (EOG) and 3) video oculography (VOG) \cite{paperno2003new}. Video or photo oculography uses image processing techniques to measure the iris position noninvasively and without contact. The majority of the commercially available eye trackers use active IR illumination. The bright pupil-dark pupil method (BP-DP) \cite{zhu2002combining} is used for finding the accurate location of the pupil. However, IR-based methods need extra hardware and specifically zoomed cameras that limit the movement of the head. Further, the accuracy of IR-based method falls drastically in uncontrolled illumination conditions. This paper proposes an image-based algorithm for localizing and tracking the eye in the visible spectrum. The main advantage of such a method is that it does not require any additional hardware and can work with regular low-cost webcams. 
Several approaches have been reported in the literature for the detection of iris centre in low-resolution images. These methods can be broadly classified into four categories 1) Model-based methods, 2) Feature-based methods, 3) Hybrid methods, and 4) Learning-based methods.
Model-based approaches generally approximate iris as a circle. The accuracy of such methods may fall when model assumptions are violated. In feature-based methods \cite{hansen2010eye}, local features like gradient information, pixel values, corners, isophote properties, etc. are used for the localization of IC. Hybrid methods combine both local and global information for higher accuracy than individual methods alone. Learning-based methods \cite{zhang2015appearance} try to learn representations from labelled data rather than using the heuristic assumptions. 
 A hybrid approach for the detection and tracking of iris centre accurately in low-resolution images is presented here. A two-stage algorithm is proposed for localizing the iris centre. A novel convolution operator is derived from Circular Hough transform (CHT) for IC localization. The new operator is efficient in the detection of IC even in partially occluded conditions and extreme corner positions. Additionally, an edge-based refinement and ellipse fitting are carried out to estimate the IC parameters accurately. IC and eye corners are used in a regression framework to determine the Point of Gaze (PoG). 
The important contributions of the paper are:
\begin{itemize}
\item A novel hybrid convolution operator for the fast localization of iris centre
\item An efficient algorithm that can estimate the iris boundary in low-resolution grayscale images
\item A framework for the eye gaze tracking in low-resolution image sequences.
\end{itemize}

The rest of the paper is organized as follows. Section 2 presents representative methods available in the literature for IC localization and gaze tracking in low resolution images. Section 3 presents the proposed algorithm. Section 4 discusses the experiments and results, and Section 5 concludes the paper.

\section{Related works}

The localization of iris or pupil is an important stage in gaze tracking. Once the iris centre has been successfully localized, regression-based methods can be used for finding the corresponding gaze points on the screen. Most of the passive image-based methods treats iris localization as a circle detection problem. Circular Hough Transform (CHT) is a standard method used for detection of circles \cite{illingworth1988survey}. Young \textit{et al}. \cite{young1995specialised} reported a method for the detection of iris with specialized Hough transform and tracking with active contour algorithm. However, this method requires high-quality images obtained from a head mounted camera.\\

Smereka \textit{et al.} \cite{smereka2008circular} presented a modified method for the detection of circular objects. They used the votes from each sector along with the gradient direction to detect circle locations. Atherton \textit{et al.} \cite{atherton1999size} proposed phase combined orientation annulus (PCOA) method for the detection of circles with convolutional operators. The annulus is convolved with the edge image to detect the peaks. Peng Yang \textit{et al.} \cite{yang2004novel} presented an algorithm for first localizing the eye region with Gabor filters and then localizing the pupil with a radial symmetry measure. However, the accuracy of the method falls when the iris moves to corners. Valenti \textit{et al.} proposed \cite{valenti2012accurate} an isophote property based iris centre localization algorithm. The illumination invariance of isophote curves along with gradient voting is used for the accurate detection of iris centres. This method is further extended in \cite{valenti2012combining} for scale invariance using scale-space pyramids. The face pose and iris centre obtained are combined to determine the point of gaze (PoG) achieving an average accuracy of 2-5 degrees in unconstrained situations. The accuracy of the method falls when iris moves towards the corners resulting in false detection of eyebrows and eye corners as iris centres. Timm \textit{et al.} \cite{timm2011accurate} proposed a method using gradients of the eye region. An extensive search is carried out in all pixels maximizing the inner product of the normalized gradient and normalized distance vector. IC is obtained as the maximum of weighted function in the region of interest. The time taken for search increases with increase in the search area. The performance of the algorithm degrades in noisy and low-resolution images where the edge detection method fails.\\

D'Orazio \textit{et al.} \cite{d2002ball} have reported a method for detection of iris centre using convolution kernels. The kernels are convolved with the gradient of images and peak points are selected as candidate points. The mean absolute error similarity measure is used to reject false positive cases. Daugman \cite{daugman2004iris} proposed an integro differential operator (IDO) for the accurate localization of iris in IR images. Curve integral of gradient magnitudes is computed to extract the iris boundary. Recently Baek \textit{et al.} \cite{baek2013eyeball} presented an eyeball model based method for gaze tracking. Elliptical shapes for eye model is saved in the database and used at the detection time for finding the iris centres. A combined IDO and a weighted combination of features are used for the localization of iris centre. Polynomial regression methods were used for training the system. They obtained average accuracy of 2.42 degrees visual angles. Sewell and Komogortsev \cite{sewell2010real} developed an artificial neural network based method for gaze estimation from low-resolution webcam images. They trained the neural network directly with the pixel values of the detected eye region. They obtained an average accuracy of 3.68 degrees. Zhou \textit{et al.} \cite{zhou2004projection} proposed a generalized projection function (GPF) that uses various projection functions and a special case hybrid projection function in localizing the iris centre. The peak positions of vertical and horizontal GPF are used to localize the eye. Bhaskar \textit{et al.} \cite{bhaskar2003blink} proposed a method for identifying and tracking blinks in video sequences. Candidate eye regions are identified using frame differencing and are subsequently tracked using optical flow.  The direction and magnitude of the flow are used to determine the presence of blinks. They obtained an accuracy of 97\% in blink detection. Wang \textit{et al.} \cite{wang2003eye} proposed one circle method where the detected iris boundary contours are fitted with an ellipse and back projected to find the gaze points. Recently many learning based methods has been proposed for iris centre localization and gaze tracking. Markuš \textit{et al.} \cite{markuvs2014eye} proposed a method for localizing pupil in images using an ensemble of randomized trees. They used a standard face detector to localize face and eye regions. Ensemble of randomized trees model was trained using the eye regions and ground truth locations. Their method obtained good accuracy in BioID database. However, the accuracy of gaze estimation is not discussed in their work. Zhang \textit{et al.} \cite{zhang2015appearance} proposed an appearance based gaze estimation framework based on Convolutional Neural Network (CNN). They have trained a CNN model with a large amount of data collected in real-world conditions. Normalized face images and the head poses obtained from a face detector were used as the input to the CNN to estimate the gaze direction. They obtained good accuracy in person and pose independent scenarios. However, the accuracy for person dependent case is lower than current geometric model based methods. The accuracy may increase with larger amount of training data, but the time taken for on-line data collection and training becomes prohibitive. Schneider \textit{et al.} \cite{schneider2014manifold} proposed a manifold alignment based method for appearance based person independent gaze estimation.  From the registered eye images, a wide variety of feature extraction methods like LBP histogram, HoG, mHoG and DCT coefficients were extracted. A combination of LBP and mHOG based features obtained the best performance. Several regression methods were used for appearance based gaze estimation. Sub manifolds for each individual were obtained using the ground truth gaze locations. Synchronized Delaunay Sub manifold Embedding (SDSE) method was used to align the manifolds of different persons. Even though their method achieved better performance compared to other appearance-based regression methods, the effect of head pose variations on the accuracy was not discussed.
Sugano \textit{et al.} \cite{sugano2014learning} proposed a person and head pose independent method for appearance based gaze estimation. They captured the images of different persons using a calibrated camera, and images corresponding to various head poses were synthesized.  An extension of random forest algorithm was used for training. The appearance of eye region and the head pose is used as the input to the algorithm which learns a mapping to the 3D gaze direction. \\
  Most of the methods proposed in literature fail when iris moves towards the corners. Another problem is regarding eye blinks, most of the algorithms returns false positives when the eyes are closed. A stable reference point is required along with the IC location for PoG estimation. Learning-based methods require large amounts of labelled data for satisfactory performance. The performance of such methods deteriorates when imaging conditions are different. Training for person dependent models require large amounts of data and often require a considerable amount of time. This limits the deployment of such methods in mobiles, tablets, etc.
In the proposed method, IC can be accurately localized even in extreme corner locations using the ellipse approximation. The high computational complexity is avoided using the two-stage scheme. An eye closure detection stage is added to prevent false positives. The localization error can be minimized by tracking the IC in the subsequent frames. The estimated IC is used in a regression framework to estimate the PoG.

\section{Proposed algorithm}
Different stages of the proposed framework are described here. 
\subsection{ Face detection and eye region localization}
Knowledge of the position and pose of the face is an essential factor in determining the point of gaze. Detection and tracking of the face help in obtaining candidate regions for eye detection. This reduces the false positive rate as well as computation time. Haar-like feature based method \cite{viola2001rapid} is used for face detection because of its higher accuracy and faster execution. An improved implementation of face detection and tracking has been proposed in our earlier work \cite{dasgupta2013vision}. The modified algorithm can detect in-plane rotated images with an affine transform based algorithm. Processing is carried out in the down sampled images to make the detection faster. The search space of detector algorithm is dynamically constrained based on the temporal information, which further increases face detection speed. Kalman filter-based tracking is used to remove the false detections and to predict the location of the face when it is not detected. The de-rotated eye region obtained is used in subsequent stages, which makes the performance of the algorithm invariant to in-plane rotations. The purpose of the de-rotation stage is only to provide a de-rotated ROI for the further processing stages. The accuracy of face rotation estimation in the pre-processing stage is only up to ±15 degrees. More accurate in-plane face rotation is obtained in the later stage using the angle of the line connecting the inner eye corners. With the improved face-tracking scheme, the frame rates of processing increase greatly (up to 200 frames per second). The analysis and tradeoffs of the algorithm are presented in our earlier work \cite{dasgupta2013vision}.\\

\subsection{ Iris centre localization}
\begin{figure*}[!htb]
\centering
\includegraphics[width=0.9\linewidth]{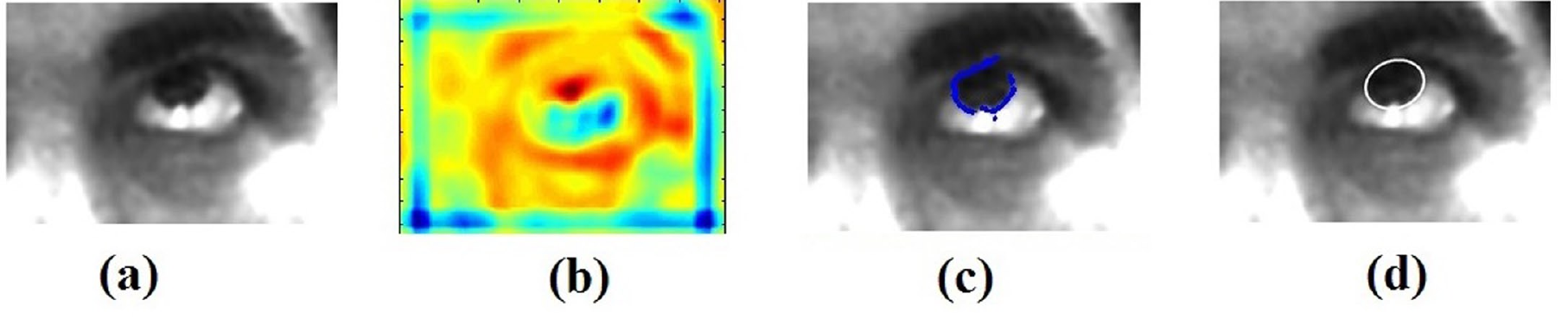}
\caption{Stages in ellipse fitting:  (a) Cropped eye region, 
 (b) Correlation surface from the proposed operator,
 (c) Selected candidate boundary points,
 (d) Fitted ellipse.}
\label{fig:1}
\end{figure*}

The proposed method uses a coarse to fine approach for detecting the accurate centre of the iris. The two-stage approach reduces the computational complexity of the algorithm as well as reduces false detection. The stages in IC localization are shown in Fig. \ref{fig:1}.
\subsubsection{ Coarse iris centre detection} Iris detection is formulated as circular disc detection in this stage. An average ratio between the width of face and iris radius was obtained empirically. For a particular image, the radius range is computed using this ratio and width of the detected face. The image gradient of iris boundary points will always be pointing outwards. The gradient directions and intensity information is used for the detection of eyes. The gradients of the image are invariant to uniform illumination changes. \\

A novel convolution operator is proposed to detect peak location corresponding to the centre of the circle. A class of convolution kernels, known as Hough Transform Filters \cite{atherton1999size} are used for this purpose. In CHT filter, the 3D accumulator is collapsed to a 2D surface by selecting a range for the radii.  \\

The 2D accumulator can be calculated efficiently using a convolution operator. A CHT filter is derived, which acts directly upon the image without any requirement of edge detection. A vector convolution kernel is designed for correlating with the gradient image, which gives a peak at the centre of the iris. \\

The convolution operator is designed as a complex operator with magnitudes as unity. The operator detects a range of circles by taking dot products with orientation inside the radius range. The equation is similar to orientation annulus proposed by Atherton \textit{et al.} \cite{atherton1999size}. The equation of convolution kernel is given as

\begin{equation}
\matrix{
   {}  \cr 
   {{O_{COA}}\left( {m,n} \right)}  \cr 
   {}  \cr 
 }  = \left\{ {\matrix{
   \matrix{
  {1 \over {\sqrt {{m^2} + {n^2}} }}\left( {\cos {\theta _{mn}} + i\sin {\theta _{mn}}} \right), \hfill \cr 
  iff,{R_{{{\min }^2}}} < {m^2} + {n^2} < {R_{{{\max }^2}}} \hfill \cr}   \cr 
   {0,otherwise}  \cr 
 } } \right.
\end{equation}

where,

\begin{equation}
{\theta _{mn}} = {\tan ^{ - 1}}\left( {{\raise0.7ex\hbox{$n$} \!\mathord{\left/
 {\vphantom {n m}}\right.\kern-\nulldelimiterspace}
\!\lower0.7ex\hbox{$m$}}} \right)
\end{equation}


Where $m$ and $n$  denote the coordinates of the kernel matrix with respect to the origin. The operator is scaled for equal contributions of circles in the radius range. A weighting matrix kernel ($W_{A}$) is also used for finding regions with maximum dark values

\begin{equation}
{W_A}\left( {m,n} \right) = \left\{ {\begin{array}{*{20}{c}}
{\frac{1}{{\sqrt {{m^2} + {n^2}} }},iff,{m^2} + {n^2} < R_{\max }^2}\\
{0,otherwise}
\end{array}} \right.
\end{equation}

the gradient complex orientation annulus can written as,\\
\begin{equation}
{C_{GCOA}} = {\mathop{\rm Re}\nolimits} \left( {{O_{COA}}} \right) \otimes {S_x} + i{\mathop{\rm Im}\nolimits} \left( {{O_{COA}}} \right) \otimes {S_y}
\end{equation}

Where $\otimes$ denotes the convolution operator; ${S_x}$ and ${S_y}$  denote the $3 \times 3$ Schaar kernels in $x$  and $y$  directions respectively. Schaar differential kernel is used owing to its mathematical properties in gradient estimation. In most of the cases, the upper portion of the iris is occluded by eyelids. An additional weighing factor ($\beta$) is included to increase the contribution of horizontal gradients. Equation for convolution kernel can be made a real-valued kernel as

\begin{equation}
{C_{RCC}} = \beta {\mathop{\rm Re}\nolimits} \left( {{O_{COA}}} \right) \otimes {S_x} + \frac{1}{\beta }{\mathop{\rm Im}\nolimits} \left( {{O_{COA}}} \right) \otimes {S_y}
\end{equation}

Where $\beta $ denotes the weighting factor.
The average intensity of each point in image can be obtained by convolving the weighting kernel with the negated version of the image as,

\begin{equation}
W = \left( {255 - I} \right) \otimes {W_A}
\end{equation}

Where $I$  and  ${W_A}$  denote the image and the kernel for computing the intensity component respectively. The final correlation output ($CO$) can be obtained by combining the convolution results for both gradient and intensity kernels as,
\begin{equation}
CO = \lambda \left( {I \otimes {C_{RCC}}} \right) + \left( {1 - \lambda } \right)W
\end{equation}

Where, $\lambda  \in [0,1]$ is a scalar, which is used to obtain the weighted combination of gradient information and image intensity to reduce spurious detections.  Iris centre corresponds to the maximum of correlation surface $CO$. Further, it is possible to represent all these operations with a single real convolution kernel, which can be applied on the image without any pre-processing, making the iris centre localization procedure even faster. For bigger circles, convolution can be carried out in Fourier domain for increasing the speed of the computation. \\

The peak of correlation output alone may lead to false detections in partially occluded images. Here, peak to side lobe ratio (PSR) of the points are used to find the iris location. The PSR values calculated in each of the local maxima and the point with maximum PSR is considered as the iris centre. The PSR is computed as:
\begin{equation}
PSR = \left( {\frac{{C{O_{\max }} - \mu }}{\sigma }} \right)
\end{equation}
Where $C{O_{\max }}$  is the local maxima in the correlation output,  $\mu $ and  $\sigma $  are the mean and standard deviation in the window around the local maxima. We have used a window size of $11 \times 11$ in this work. The point with the maximum  $PSR$ is selected as the iris centre.

\subsubsection{Sub-pixel edge refining and ellipse fitting}

In this stage, the rough centre points obtained in the previous stage are used to refine the IC location. The objective is to fit the iris boundary with an ellipse. The constraints on the major and minor axis can be obtained empirically ( $R_{min}$ and $R_{max}$  ). The algorithm presented searches in the radial direction similar to Starburst algorithm \cite{li2005starburst}. However, the search process finds only the strongest edges with similar gradients. Dominant edges with agreeing directions are selected with sub-pixel accuracy. An angle versus distance plot is obtained and the outlier points are filtered using median filter. An ellipse can be fitted to five points by the least square method using Fitzgibbon's algorithm [45]. However, we used this algorithm in a RANSAC framework for minimizing the effect of outliers.  RANSAC algorithm is employed \cite{fischler1981random} for ellipse fitting, using the gradient agreement \cite{swirski2012robust} of the detected boundary points and the fitted ellipse as the support function. Additionally, a modified goodness of fit   is evaluated as the integral of dot products of outward gradients over the detected boundary (only agreeing gradients). The parameters obtained are considered as false positives if the goodness of fit is less than a threshold. The detailed algorithm for ellipse fitting is given in Algorithm 1.
\begin{equation}
GoF = \sum\limits_{x,y \in f(\lambda )}^{} {\left( {\min \left( {\frac{{\nabla f(x,y)}}{{\left| {\nabla f(x,y)} \right|}} \bullet \nabla I(x,y),0} \right)} \right)} 
\end{equation}
\subsection{Iris tracking}
Kalman filter (KF) \cite{yoon2008new} is used to track the IC in a video sequence. The search region for iris detection can be limited with the tracking approach. Once the IC is detected with sufficient confidence, the point can be tracked in subsequent frames easily. Face detection stage can be avoided in this case. KF \cite{kiruluta1997predictive} estimates can be used as the corrected estimates for iris position.\\ 
In the current tracking application, constant velocity model is selected as the transition model. Coordinates of the centre of iris along with their velocities are used as states. \\
\begin{equation}
{X_{k + 1}} = {F_k}{X_k} + {W_k}
\end{equation}
Where, ${X_k}$ is the state containing  $x,y,{v_x},{v_y}$ (coordinates and velocities in  $x$  and $y$   directions respectively) at the ${k^{th}}$ instant.  The measurement noise covariance matrix is computed from the measurements obtained during the gaze calibration stage. The process covariance matrix is computed empirically. Measurements obtained from the IC detector are used to correct the estimated states.


\begin{figure}[!htb]
\centering
\includegraphics[width=1\linewidth]{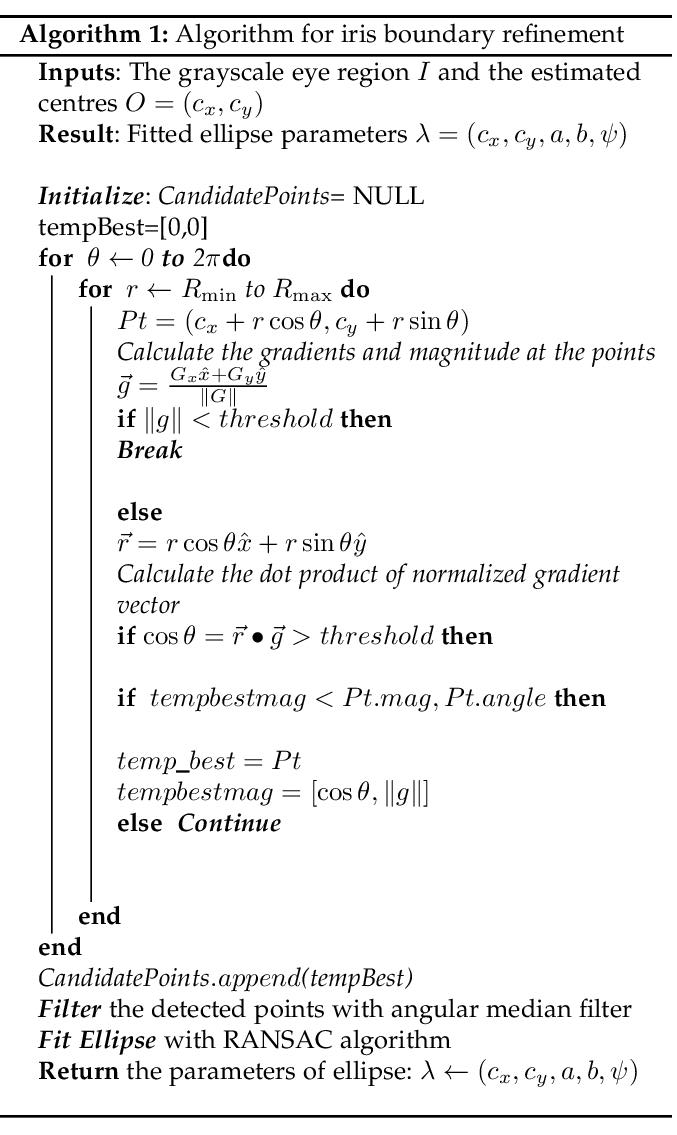}
\caption{Algorithm for iris boundary refinement.}
\label{alg}
\end{figure}

\subsection{Eye closure detection}
The IC localization algorithm may return false positives when the eyes are closed. Thresholds on the peak magnitude were used to reject false positives. However, the quality of peak may degrade in conditions such as low contrast, image noise and motion blur. The accuracy of the algorithm may fall in these conditions, and hence machine learning based approach is used to classify the eye states as open or close. The Histogram of Oriented Gradients (HOG) \cite{dalal2005histograms} features of eye regions are calculated and a Support Vector Machine (SVM) based classifier is used to predict the state of the eye. The HOG features are computed in the detected ROI for left and right eyes separately. The SVM classifier was trained offline from the database. If the eye state is classified as closed, then the predicted value from KF is used as the tentative position of the eye. If eyes are detected as open, then the result from the two-stage method is used to update the KF.
\subsection{Eye corner detection and tracking}
The appearance of inner eye corner exhibits insignificant variations with eye movements and blinks. Therefore, this paper proposes to use inner eye corners as reference points for gaze tracking. The eye corners can be located easily in the eye ROI. The vectors connecting eye corners and iris centres can be used to calculate gaze position. Several methods have been proposed in the literature for the localization of facial landmarks \cite{cristinacce2006facial}. In the proposed method Gabor jets \cite{vukadinovic2005fully} are used to find eye corners in the eye ROI owing to its high accuracy.  The detected eye corners are tracked in the subsequent frames using optical flow and Normalized cross correlation (NCC) \cite{lewis1995fast}, \cite{tomasi1991detection} based method.  The tracker is automatically reinitialized if the correlation score is less than a pre-set threshold value.
\subsection{Gaze estimation}
Gaze point can be computed from the IC location and a reference point. Earlier works \cite{pires2013visible}, \cite{sigut2011iris} have used the eye centre and corneal reflections as reference points. False detection in any of the corners will result in performance degradation of the algorithm. Hence, the inner eye corners are used as reference points in this work. Detection of the eye corner in every frame might increase the error rates and computational complexity. We avoid this issue by tracking of eye corners in the frames which ensure stable reference points. If $({x_1},{y_1})$ and $({x_2},{y_2})$   denote the coordinates of eye corner and iris centre respectively, the EC-IC vector (with reference to the corner) can be obtained as: $(x,y) = ({x_2} - {x_1},\,\,{y_2} - {y_1})$ . The eye corner–iris centre vector (EC-IC) is calculated separately for the left and right eye.
\\

\subsubsection{Calibration}
In the calibration stage, subjects were asked to look at uniformly distributed positions on the screen. The EC-IC vectors along with gaze points are recorded. The mapping between EC-IC vector and screen coordinates is nonlinear because of the angular movement of the iris. We used two different models for the mapping between EC-IC vector and point of gaze (PoG), 1) polynomial regression and 2) a radial basis function (RBF) kernel based method. In polynomial regression, a second order regression model is used for determining the point of gaze since it offers the best trade-off between model complexity and accuracy.

\begin{equation}
screen{X_i} = {a_0}{x_i} + {a_1}{y_i} + {a_2}{x_i}{y_i} + {a_3}x_i^2 + {a_4}y_i^2 + {a_5}
\end{equation}
\begin{equation}
screen{Y_i} = {b_0}{x_i} + {b_1}{y_i} + {b_2}{x_i}{y_i} + {b_3}x_i^2 + {b_4}y_i^2 + {b_5}
\end{equation}

Where, $ \left( {{x_i},{y_i}} \right)$  are the components of EC-IC vector and, $\left( {screen{X_i},screen{Y_i}} \right)$  the corresponding screen positions. The data obtained from calibration stage is used in the least square regression framework to calculate unknown parameters.\\

In the RBF kernel based method, we used non-parametric regression \cite{nadaraya1964estimating} for estimating PoG. The components of EC-IC vector are transformed into kernel space using the following expression,

\begin{equation}
k({p_i},{p_l}) = exp( - \frac{{{{\left\| {{p_i} - {p_l}} \right\|}^2}}}{{2{\sigma _k}^2}})
\end{equation}

Where, ${p_i}$ and  ${p_l}$ denote of EC-IC vector and the landmark points respectively. ${\sigma _k}$  denote the standard deviation of the RBF function. We have tested the algorithm in both $3 \times 3$ and $4 \times 4$ calibration grids. Instead of using all the samples as landmark points, we have used only one landmark per calibration point. The number of landmarks used was 9 and 16 for $3 \times 3$ and $4 \times 4$ grid respectively. For each calibration point on the grid, the landmark vector is calculated as the median of the components of EC-IC vector at the particular point. The dimension of the design matrix is reduced by the use of landmark points (since the data points are clustered around the calibration points). Regression is carried out after transforming all the points to kernel space \cite{kohn2001nonparametric} which improved the accuracy of PoG estimation. The training procedure is carried out for left and right eyes.

\subsubsection{Estimation of PoG}

\begin{figure}[!htb]
\centering
\includegraphics[width=0.9\linewidth]{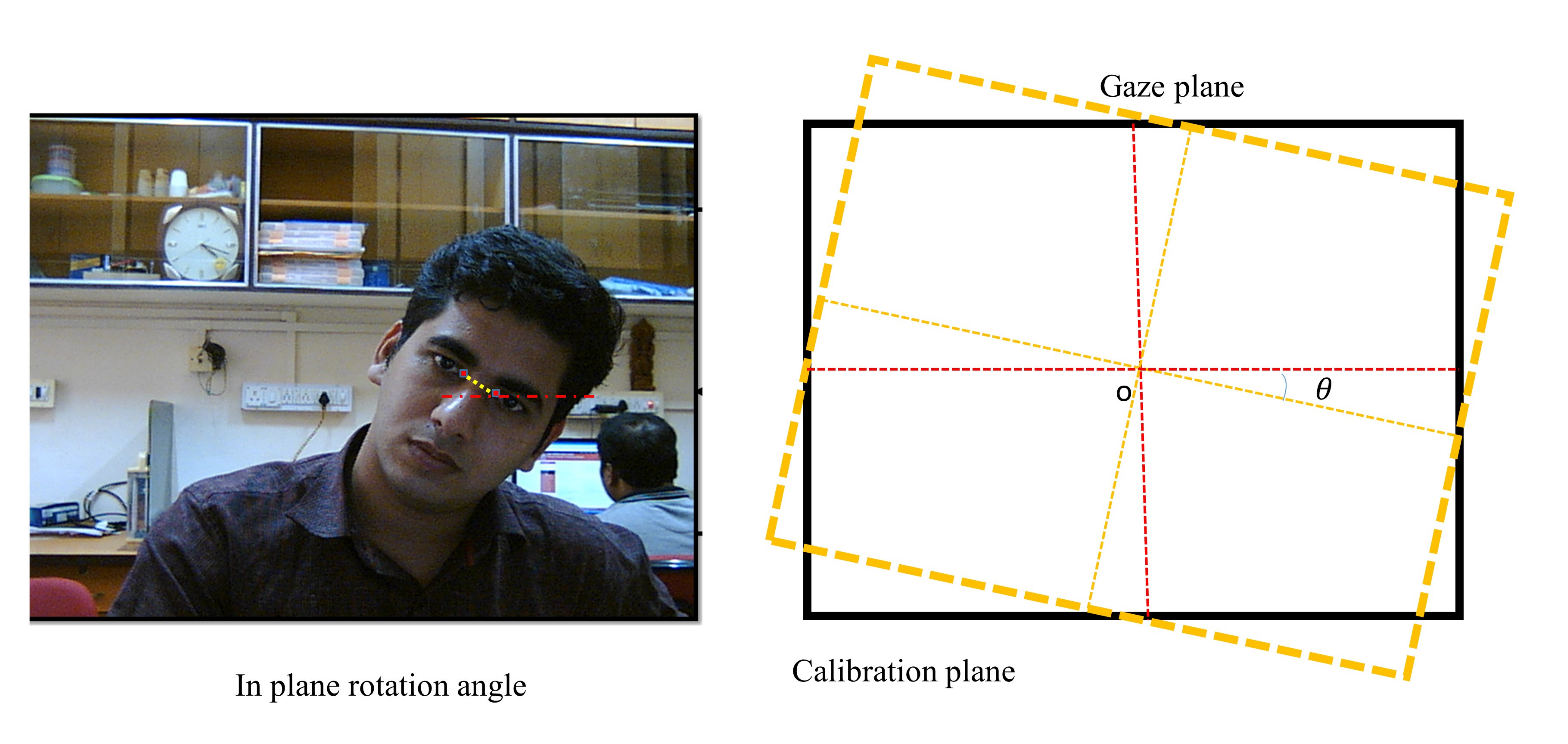}
\caption{Transformation of estimated gaze point to screen coordinates for in-plane rotation.}
\label{fig:2}
\end{figure}

The parameters obtained from calibration procedure are used to determine the gaze position. The regression function obtained is used to map the EC-IC vectors to screen coordinates. The gaze position is computed as the average position returned by the left and right eye models. The head position is assumed to be stable during the calibration stage. After calibration, the estimated gaze point will be on the calibration plane (i.e., w.r.t the position of the face during the calibration stage). Deviation from this face position would cause errors in the estimated gaze locations. The effect of 2D translation is minimal for moderate head movements since the reference points for EC-IC vector are eye corners, which also move along with face (thereby providing a stable reference invariant to 2d translational motion). Even though the method is invariant to moderate amount of translation, the accuracy falls when there is a rotation. This error can be corrected using the face pose information. The in-plane rotation of face can be calculated from the angle of the line connecting the inner corners of the left and right eye as shown in Fig. \ref{fig:2}. The rotation matrix can be computed as,

\begin{equation}
R = \left[ {\begin{array}{*{20}{c}}
{\cos \theta }&{ - \sin \theta }\\
{\sin \theta }&{\cos \theta }
\end{array}} \right]
\end{equation}

Where, $\theta $ is the difference in angle from the calibration stage. The corrected PoG can be found from coordinate transformation with the screen centre as the origin. The exact 3D pose variations can be corrected using more computationally intensive models like Active Appearance Models (AAM) \cite{cootes2001active}, Constrained Local Model (CLM) \cite{cristinacce2006feature}, etc.

\section{Experiments}

We have conducted several experiments to evaluate the performance of the proposed algorithm. The algorithm has been evaluated using standard databases and a custom database.  The IC localization accuracy is evaluated in standard databases and compared with the state of the art methods. The accuracy in PoG estimation and eye closure detection is assessed in the custom dataset.

\subsection{Experiments on IC localization}

\subsubsection{Evaluation method}
Face detection is carried out using Viola-Jones method \cite{viola2004robust}. The eye regions are localized based on anthropometric ratios. \\

The normalized error is used as the metric for comparison with other algorithms. The normalized measure for worst eye characteristics (WEC) \cite{jesorsky2001robust} is defined as: \\

\begin{equation}
{e_{WEC}} = \frac{{\max ({d_l},{d_r})}}{w}
\end{equation}

Where ${d_l}$ and  ${d_r}$ are the Euclidean distances between ground truth and detected iris centres (in pixels) of left and right eye respectively, and    is the true distance between the eyes in pixels. The average (AEC) and best of eye detection (BEC) errors are also calculated for comparison. They are defined as:

\begin{equation}
{e_{AEC}} = \frac{{({d_l} + {d_r})}}{{2w}},{e_{BEC}} = \frac{{\min ({d_l},{d_r})}}{w}
\end{equation}

Where, ${e_{BEC}}$ is the minimum error in both the eyes and ${e_{AEC}}$ is the average error of both the eyes.\\

\subsubsection{Experiments in BioID and Gi4E Database}
A comparison of the proposed method with the state of the art methods is carried out for BioID \cite{BioID} and Gi4E \cite{ponz2012dataset} databases. The BioID database consists of images of 23 individuals taken at different times of the day. The size, position and pose of the faces change in the image sequences. The contrast is very low in some images. In some images, eyes are closed. There are images where subject wear glasses and glints are present due to illumination variations. The database contains a total of 1,521 images with a resolution of $384 \times 288$ pixels. The ground truth files for left and right iris centres are also available.\\

Gi4E dataset consists of 1380 colour images of 103 subjects with a resolution of $800 \times 600$. It contains sequences where the subjects are asked to look at 12 different points on the screen. All the images are captured at indoor conditions at varying illumination levels and different backgrounds. The database represents realistic conditions during gaze tracking, head movements, illumination changes and movement of eyes towards corners and occlusions with eyelids. The ground truth of left and right eye positions is also available with the database.

\begin{figure*}[!htb]
\centering
\includegraphics[width=0.9\linewidth]{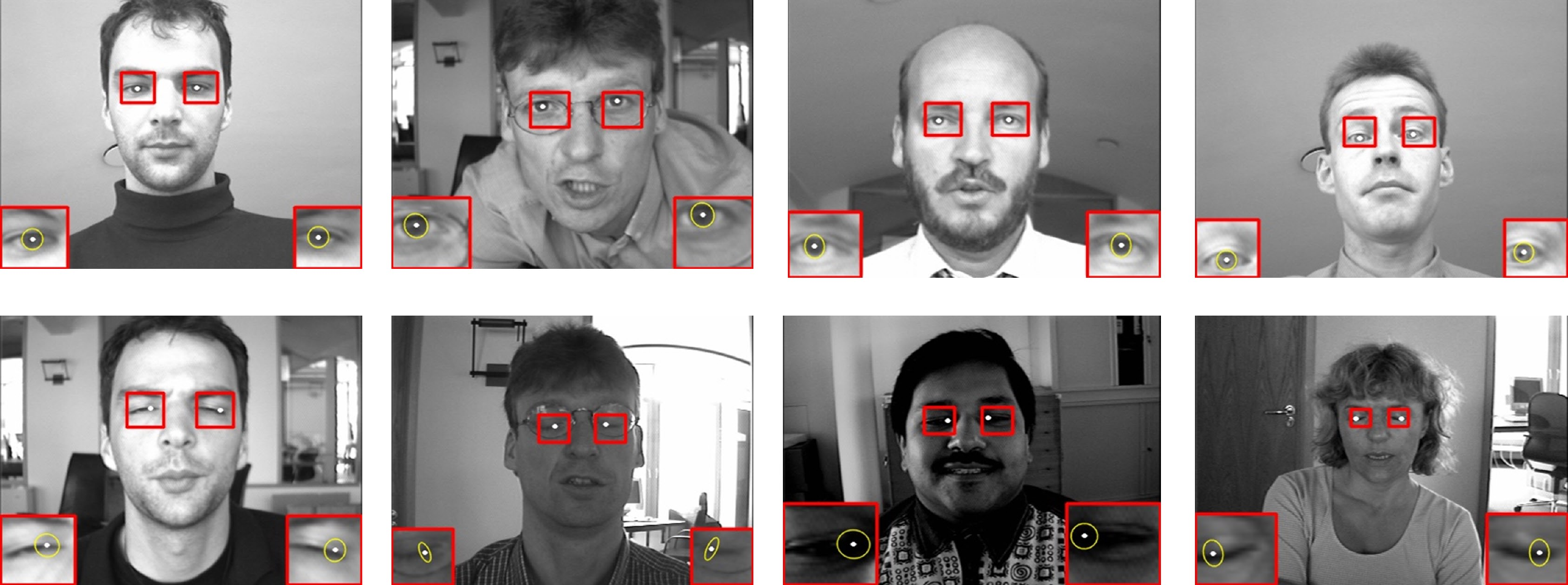}
\caption{Few samples showing successful detections (first row) and failures (second row) in BioID Database.}
\label{fig:3}
\end{figure*}

\begin{figure*}[!htb]
\centering
\includegraphics[width=0.9\linewidth]{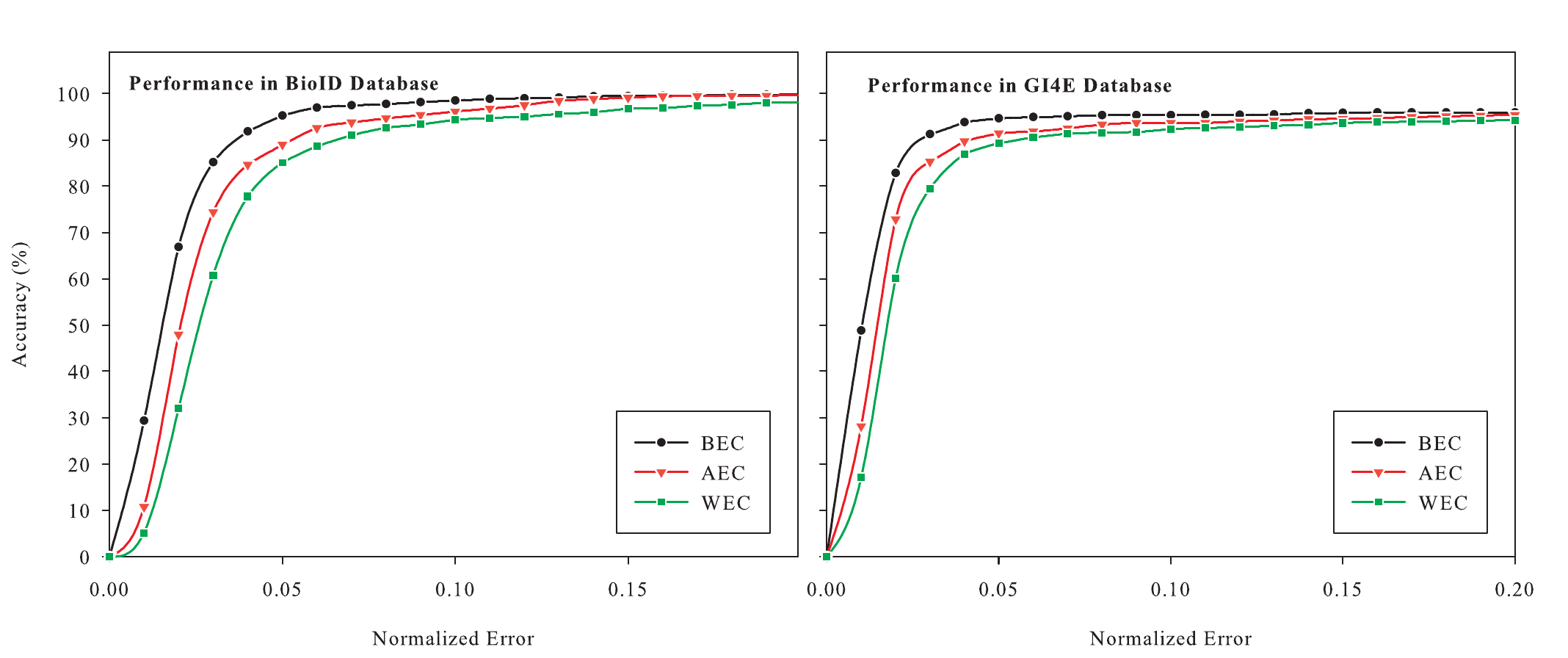}
\caption{Performance of the proposed algorithm in BioID and Gi4E databases. The graph shows three normalized measures corresponding to WEC - Worst eye characteristics, AEC-Average eye characteristics and BEC- Best Eye characteristics.}
\label{fig:4}
\end{figure*}

Fig. \ref{fig:3} show some of the correct detections and failures of the algorithm in BioID database. The face detection accuracy obtained was 94.74\%. In most of the cases, errors are due to partial closure of eyes and eyeglasses. The algorithm performs well when eyes are visible even with low contrast and varying illumination levels. Fig. \ref{fig:4} show the performance of proposed algorithm in BioID and Gi4E database. The value of  $\lambda $and $\beta $  used were 0.95 and 2 respectively. The proposed algorithm gives an accuracy (WEC) of 85.08 for $e \le 0.05$. 


\begin{figure*}[!htb]
\centering
\includegraphics[width=0.9\linewidth]{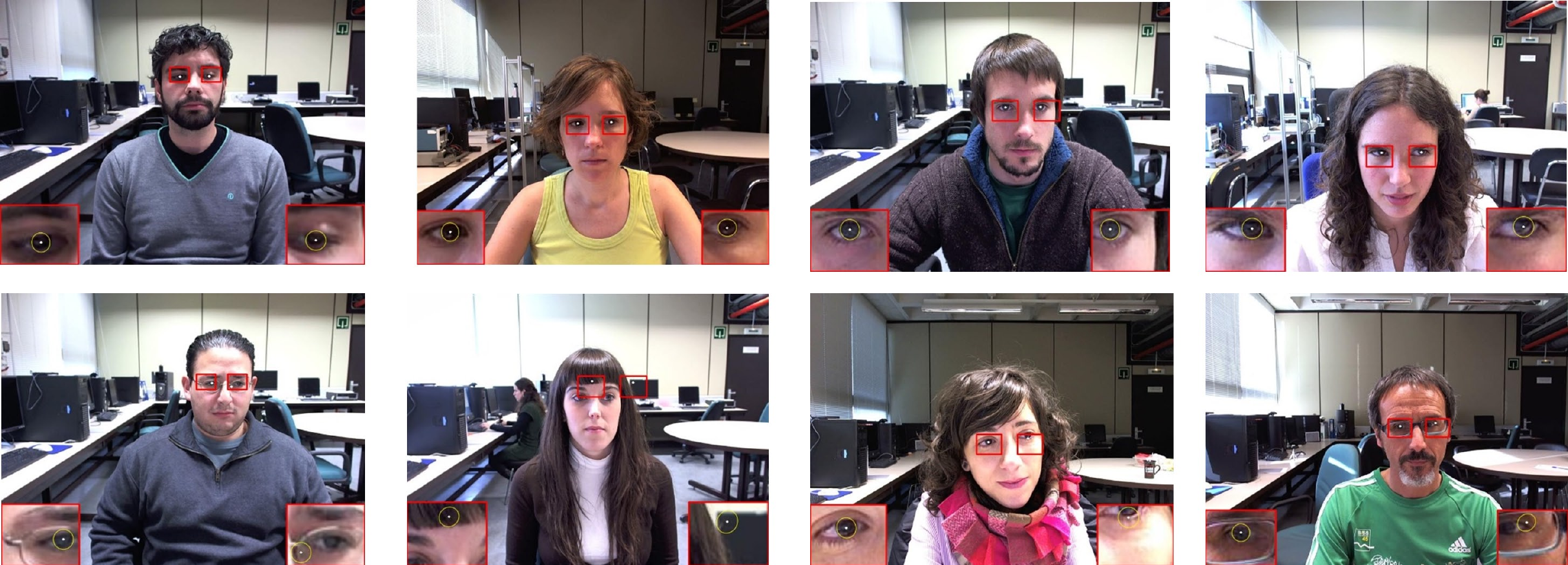}
\caption{Some samples showing successful detections (first row) and failures (second row) in Gi4E Database.}
\label{fig:5}
\end{figure*}

In Gi4E database, the worst-case accuracy (WEC) is 89.28 for $e \le 0.05$ . Fig. \ref{fig:5} show results from the algorithm. The face detection accuracy obtained was 96.95\%.  The main advantage is that the algorithm performs well in different eye gazes positions which is essential in gaze tracking applications.


\begin{figure*}[!htb]
\centering
\includegraphics[width=0.9\linewidth]{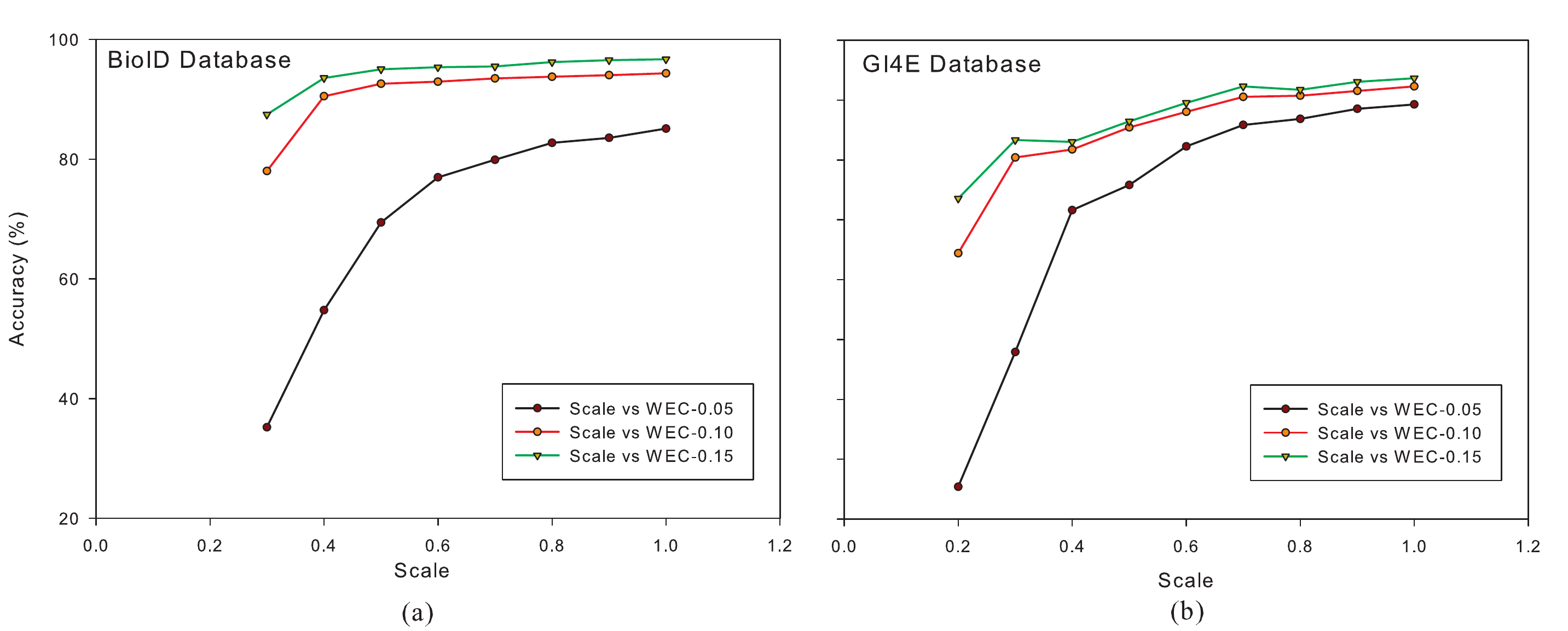}
\caption{WEC Performance of the proposed algorithm in (a) BioID and (b) Gi4E databases with different resolutions. Scaling parameter is w.r.t the original image resolutions in the corresponding databases. }
\label{fig:6}
\end{figure*}

The performance of the algorithm may vary depending upon the distance of the user from the monitor. This effect is emulated using images with different spatial resolutions. The performance of the proposed algorithm in different spatial resolutions in BioID and Gi4E database is shown in Fig. \ref{fig:6}. The accuracy of iris localization falls as the image resolution decreases. However, the detection accuracy (WEC-0.5) is more than 80\% for scaling up to 0.8 ($307 \times 230$ resolution) and 0.6 ($480 \times 360$ resolution) in BioID (82.72\%) and Gi4E (82.24\%) databases respectively. 

\subsubsection{Comparison with state of the art methods}

\begin{figure*}[!htb]
\centering
\includegraphics[width=0.9\linewidth]{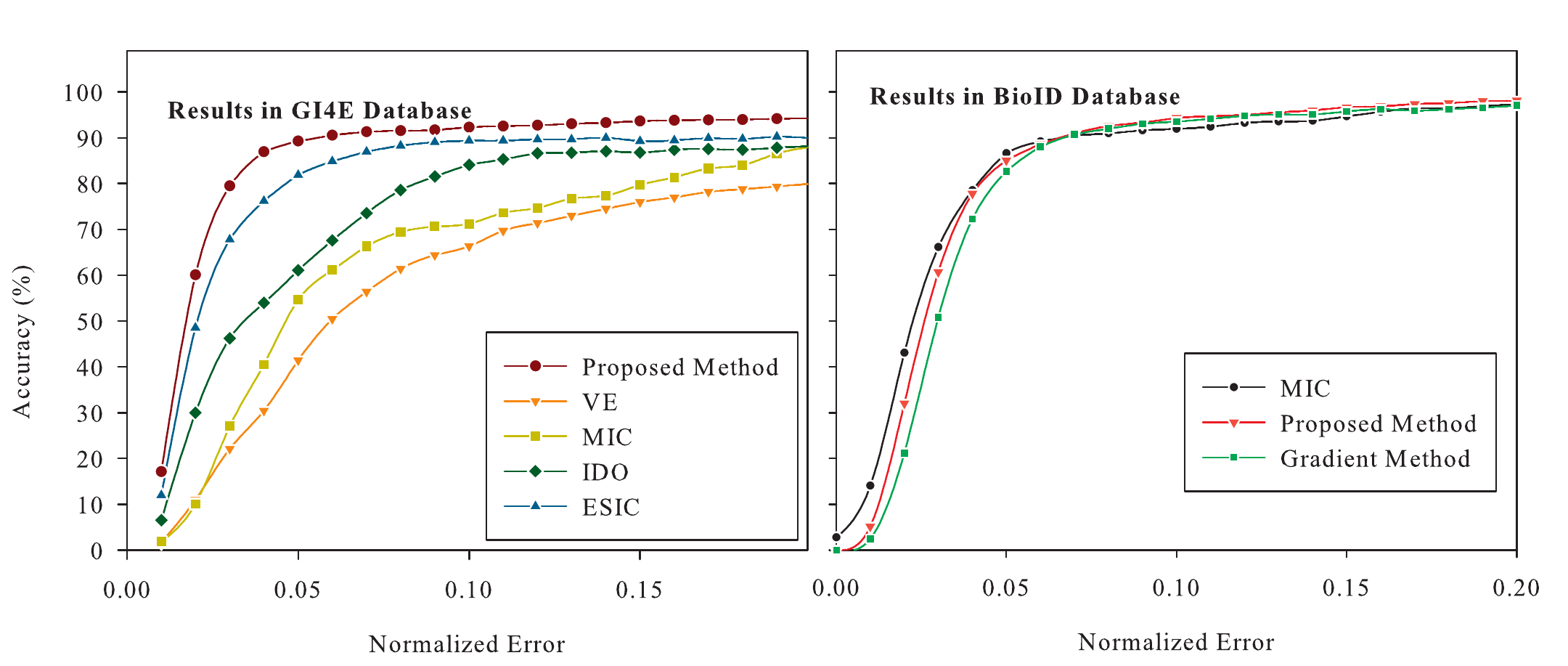}
\caption{ WEC performance comparison of proposed method with state of the art methods in Gi4E and BioID databases. }
\label{fig:7}
\end{figure*}

We have compared the algorithm with many state of the art algorithms in BioID and Gi4E databases. The algorithms MIC \cite{valenti2012accurate}, Timm \textit{et al.} \cite{timm2011accurate}  and the proposed methods are tested in BioID database. The evaluation is carried out with normalized worst eye characteristics (WEC). The results are shown in Fig. \ref{fig:7}. The WEC data is taken from ROC curves given in author’s papers. The algorithm proposed is the second best in BioID database as shown in Table \ref{tab:2}. Isophote method (MIC) performs well in this database. The proposed algorithm fails to detect accurate positions when eyes are partially or fully closed (eye closure detection stage was not used here). Presence of glints is another major problem. The failure of face detection stage and reflections from the glasses causes false detections in some cases. The addition of a machine learning based classification of local maxima may improve the results of proposed algorithm.

Gi4E is a more realistic database for eye tracking purposes. It contains images with head and eye movements. The algorithms for comparison are chosen as VE \cite{wang2003eye}, IDO \cite{daugman2004iris}, MIC \cite{valenti2012accurate}, ESIC \cite{baek2013eyeball}. The results are compared with WEC values obtained from ROC curves reported in Baek \textit{et al.} \cite{baek2013eyeball}. It is seen (Table. \ref{tab:1}) that the proposed method outperforms all. The accuracy of MIC method is very low when the eyes move to the corners. The circle approximation of most of the algorithms fails when eyes move towards the corners, making them inapt for eye gaze tracking applications. The performance evaluation is carried out on each frame separately. Addition of temporal information described in section 3.3 might increase the accuracy of the algorithm greatly.

\begin{table}[]
\centering
\caption{Comparison of proposed method with state of the art algorithms in Gi4E databases
Method	 }
\label{tab:1}
\noindent\begin{tabular*}{\columnwidth}{@{\extracolsep{\stretch{1}}}*{7}{r}@{}}
\toprule
\multicolumn{1}{c}{Method} & $e \le 0.05$     & $e \le 0.10$    & $e \le 0.15$     & $e \le 0.20$      \\ \midrule
\textbf{Proposed}          & \textbf{89.28} & \textbf{92.3} & \textbf{93.64} & \textbf{94.22}  \\
VE                         & 41.4  & 66.3 & 75.9  & 80*    \\
MIC                        & 54.5  & 71.2 & 79.7  & 88.1*  \\
IDO                        & 61.1  & 84.1 & 86.7  & 88.15* \\
ESIC                       & 81.4  & 89.3 & 89.2  & 89.9*  \\ \bottomrule
\end{tabular*}
\end{table}


\begin{table}[]
\centering
\caption{Comparison of proposed method with state of the art algorithms in BioID database
Method	 }
\label{tab:2}
\noindent\begin{tabular*}{\columnwidth}{@{\extracolsep{\stretch{1}}}*{7}{r}@{}}
\toprule
\multicolumn{1}{c}{\textbf{Method}} & $e \le 0.05$     & $e \le 0.10$    & $e \le 0.15$     & $e \le 0.20$      \\ \midrule
MIC(\cite{valenti2012accurate}) and Sift kNN                & \textbf{86.09} & 91.67 & 94.5\* & 96.9\* \\
\textbf{Proposed }                           & 85.08 & \textbf{94.3}  & \textbf{96.67} & \textbf{98.13} \\
Timm \textit{et al.},(\cite{timm2011accurate})                & 82.5  & 93.4  & 95.2  & 96.4  \\ \bottomrule
\end{tabular*}
\end{table}

\begin{figure}[!htb]
\centering
\includegraphics[width=0.9\linewidth]{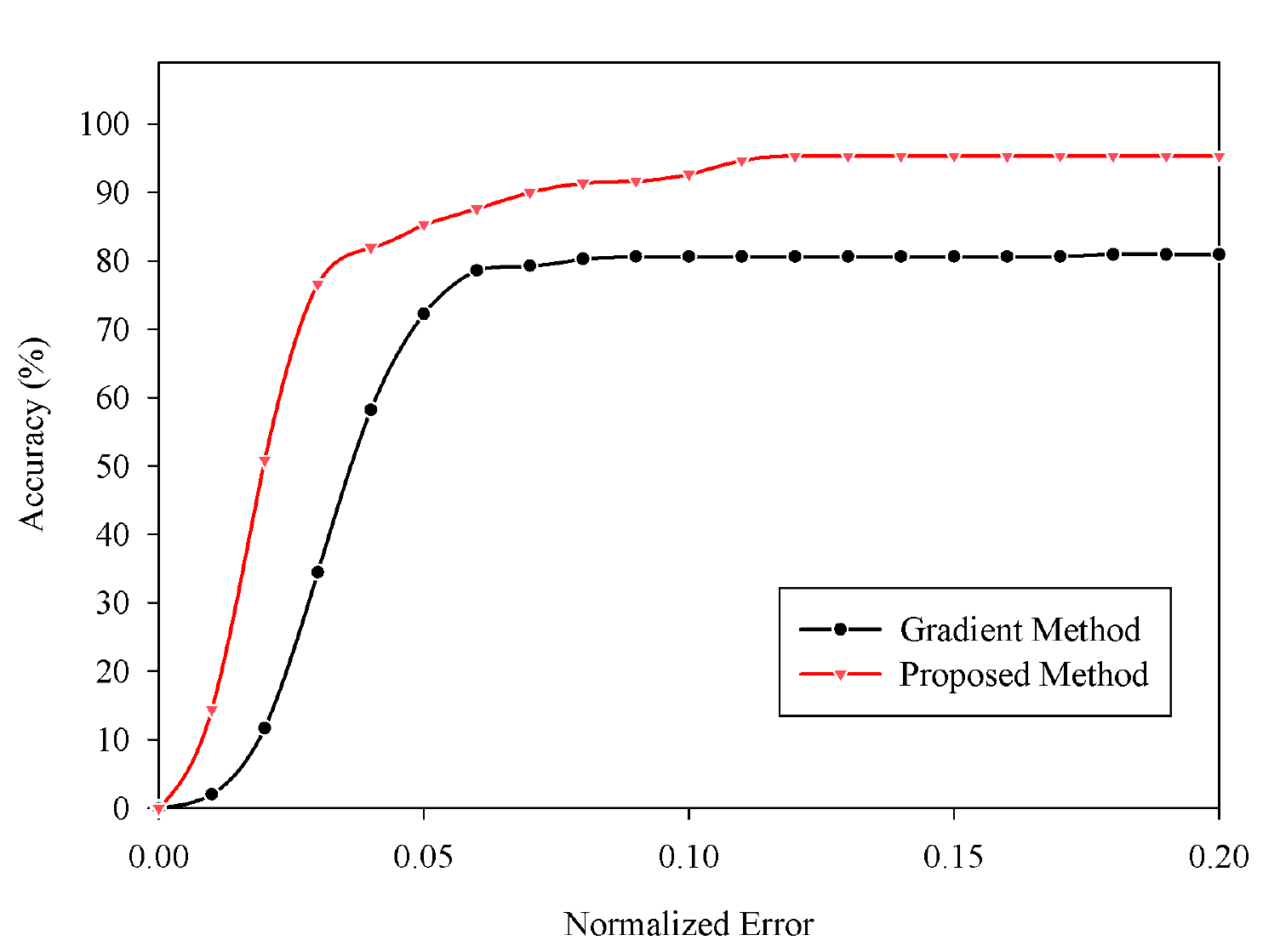}
\caption{WEC performance comparison of the proposed method with gradient based method in extreme corner cases. }
\label{fig:8}
\end{figure}

We have performed additional experiments on the Gi4E dataset to evaluate the performance when the iris moves to the corner. A subset of 299 images was selected according to the position of iris centre about the eye corner. We have compared the results with the gradient-based method for evaluating the accuracy with circle model and ellipse model. The WEC characteristics comparison is shown in Fig. \ref{fig:8}. The ellipse approximation improves the accuracy significantly compared to the circle approximation.


\subsection{Experiment with our database}
\subsubsection{Experiment for gaze estimation accuracy evaluation}

\begin{figure*}[!htb]
\centering
\includegraphics[width=0.9\linewidth]{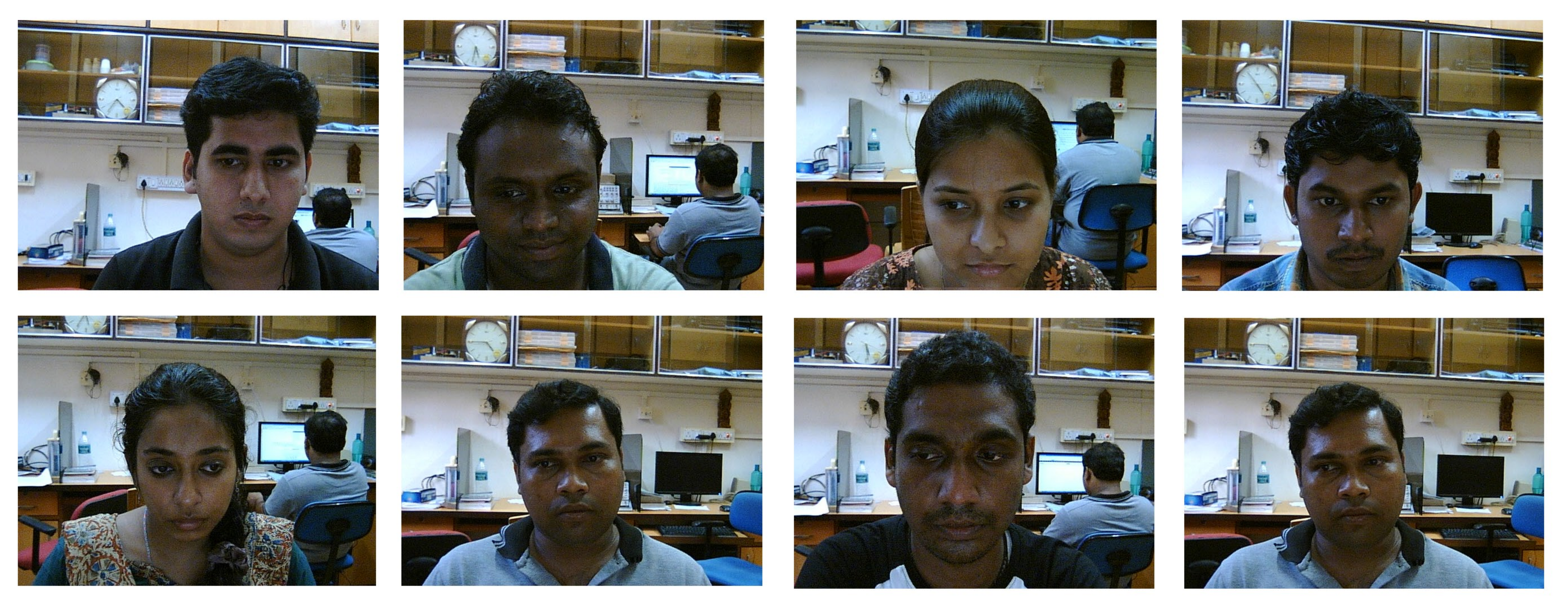}
\caption{ Sample images of subjects participated in the experiment. }
\label{fig:9}
\end{figure*}

An experiment was performed on ten subjects using a standard webcam and a 15.6- inch monitor with a resolution of $1366 \times 768$. The subjects were seated 60 cm from the screen and asked to follow the red dot on the monitor. The videos of the eye movements were recorded at 30 fps with a resolution of $640 \times 480$. The subjects were asked to look at the calibration patterns two times. We used both 9 point and 16 point calibration and compared the results. Fig. \ref{fig:9} shows some of the images from the dataset.\\

We have evaluated the IC localization accuracy on a subset of images in the in-house dataset. A subset of 1000 images was selected, and the IC localization accuracy was evaluated. The proposed approach obtained WEC accuracies of 90.2\% and 92.9\% for $e \le 0.05$  and  $e \le 0.10$ respectively.\\

 For $3 \times 3$ and $4 \times 4$ calibration grids, we have tested with polynomial regression and kernel space-based methods. The samples from the first session were used in the training stage. The parameters for regression were found from the training data. In the testing stage, the samples in the second session were used to estimate the PoG. The mean position computed from the left and right eye PoG is used as the final gaze point.  The error in the estimation is computed using the ground truth. The mean absolute error in visual angles in horizontal, vertical and overall accuracy is computed using the head distance from the screen.
 
 \begin{equation}
Accuracy = {\tan ^{ - 1}}\left( {{\raise0.7ex\hbox{${error}$} \!\mathord{\left/
 {\vphantom {{error} {head\,distance}}}\right.\kern-\nulldelimiterspace}
\!\lower0.7ex\hbox{${head\,distance}$}}} \right)
 \end{equation}

\begin{figure*}[!htb]
\centering
\includegraphics[width=0.8\linewidth]{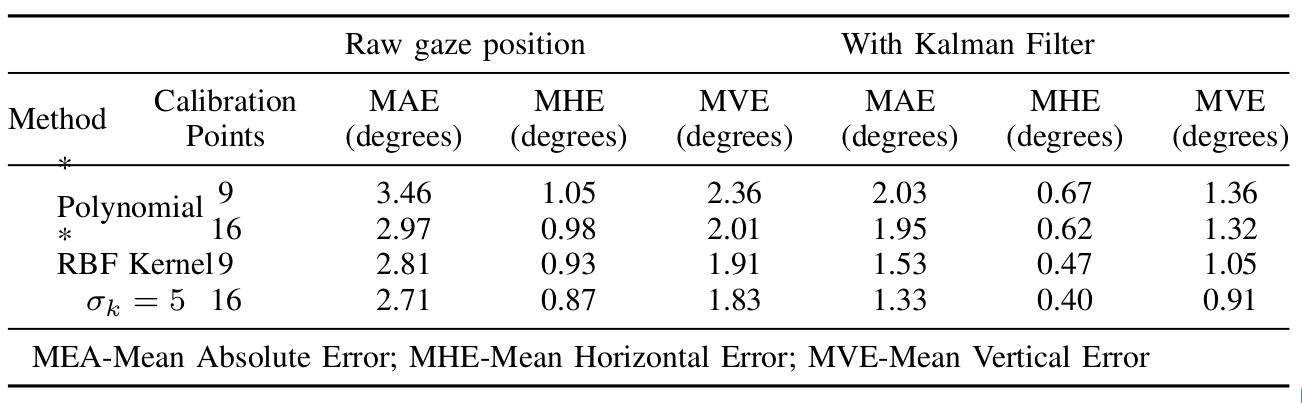}
\caption{Gaze estimation accuracy}
\label{tab:3}
\end{figure*}

\begin{figure*}[!htb]
\centering
\includegraphics[width=0.9\linewidth]{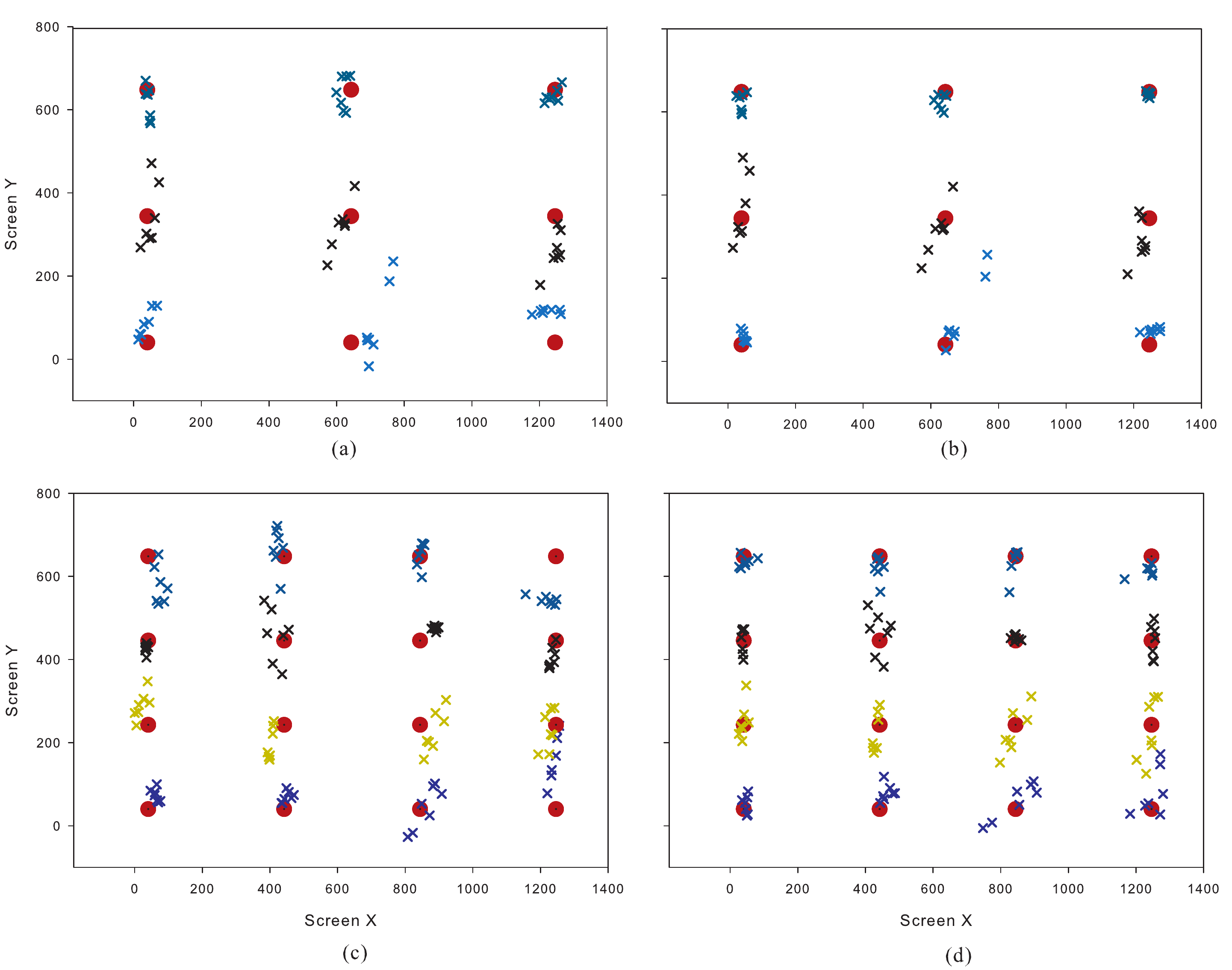}
\caption{ PoG estimates with 16 and 9 point calibration grids (a) (c) polynomial regression (b) (d) RBF kernel. Dots and crosses denote the target points and estimated gaze positions respectively. }
\label{fig:10}
\end{figure*}

The average errors are high when the EC-IC vectors are computed on a frame by frame basis. We further computed the PoG using KF estimates which reduced the jitter significantly. The results with and without  KF on $3 \times 3$ and $4 \times 4$ calibration grids are  tabulated in Table \ref{tab:3}. The qualitative results of gaze estimation stage are shown in Fig. \ref{fig:10}.

\subsubsection{Experiment for eye closure detection}
The eye regions obtained from the face detection stage are histogram equalized and resized to the size of $30 \times 30$. A data set of 4000 images containing 2000 samples for open and 2000 samples for closed eyes were formed from our dataset. HOG features were extracted from various pixel per cell windows and eight orientations. The extracted HOG features were used to train the SVM classifier. Ten times ten-fold cross validation was used to examine the accuracy of the trained classifiers. The results show that the proposed method achieves an average accuracy of 98.6\% with linear SVM. The results obtained are shown in Table \ref{tab:4}. 


\begin{table}[!htb,width=1\linewidth]
\centering
\caption{Accuracy of eye closure detection}
\label{tab:4}
\noindent\begin{tabular*}{\columnwidth}{@{\extracolsep{\stretch{1}}}*{7}{r}@{}}
\toprule
Pixel per cell in HOG & RBF Kernel SVM    & Linear SVM        \\ \midrule
2                     & 97.5\% (SD=3.1\%) & 98.3\% (SD=3.2\%) \\
4                     & 97.2\% (SD=2.7\%) & 98.6\% (SD=2.4\%) \\ \bottomrule
\end{tabular*}
\end{table}
 
\subsection{Discussions}
The proposed method contains cascaded stages of many algorithms. The gaze estimation accuracy is a good proxy for the combined accuracy of all the cascaded stages. Face and IC are tracked using Kalman filters independently due to their distinct dynamics. The tracking based framework increases the robustness by reducing the effect of per-frame localization errors.\\

For successful eye tracking using webcams, the normalized error should be less than 0.05. The proposed algorithm performs better in realistic conditions for webcam-based gaze tracking. The accuracy of gaze estimation was evaluated with the proposed approach in both $3 \times 3$ and $4 \times 4$ calibration grids. RBF kernel-based nonparametric regression method was found to perform better than second order polynomial models. The average error rate obtained with the per-frame based detection was 2.71 degrees. The accuracy of the gaze tracking improved significantly by the use of KF, which uses the temporal information effectively to reduce the error rate to 1.33 degrees. \\
One of the advantages of the proposed algorithm is low computational complexity. The eye detection, being a convolution-based method can be implemented in Fourier domain \cite{burrus1991dft} for faster computation. Multi-resolution convolution can be used to reduce the search space even further.  The algorithm was implemented in a 2.5 GHz core 2 duo PC with 2 GB RAM. C++ implementation using OpenCV library \cite{bradski2000opencv} (without multi-threading) was used for the evaluation experiments in Ubuntu 14.04 OS (32 bit) environment. It detects the face and eye corners in the first frame and tracks the eye corners over time. The temporal information is used to reduce the search space for face detection using a Kalman filter. The images were acquired using a 60 fps $640 \times 480$ webcam. The online processing speed was limited only by the lower frame rate of the camera. The offline processing speed of the entire algorithm is well over 100 fps on the recorded video. This is suitable for normal PC based implementation with 30 fps webcams. The proposed method can also be implemented in smart devices like mobile phones and tablets due to its low computational overhead. The low computational complexity makes it possible to extend the pose tracking with more complex 3D models, which could make the PoG estimation invariant to out of plane rotations as well.

\section{Conclusions}

This paper describes an algorithm for a fast and accurate localization of iris centre position in low-resolution grayscale images. A framework for tracking faces in video at very high frame rates, well above 100 fps is also presented. A two-stage iris localization is carried out, and the filtered candidate iris boundary points are used to fit an ellipse using a gradient aware RANSAC algorithm. The proposed algorithm is compared with the state of the art methods and found to outperform edge-based methods in low-resolution images. The computational complexity of the algorithm is very less since it uses a convolution operator for iris centre localization. The paper also proposes and implements a gaze-tracking framework. Inner eye corners are used as the reference for calculating gaze vector. Kalman filter based tracking is used to estimate the gaze accurately in video. Further, ellipse parameters obtained from the algorithm can be combined with geometrical models for higher accuracy in gaze tracking. We have considered only in-plane rotations in this work. However, pose invariant models can be developed by using more computationally complex 3D models.
\section{Acknowledgements}
The authors would like to thank the subjects for participation in the experiment. 
\ifCLASSOPTIONcaptionsoff
  \newpage
\fi
\bibliographystyle{IEEEtran}

\bibliography{refsiet}

\end{document}